\pdfoutput=1

\documentclass[11pt]{article}

\usepackage{EMNLP2023}

\usepackage{times}
\usepackage{latexsym}

\usepackage[T1]{fontenc}

\usepackage[utf8]{inputenc}

\usepackage{microtype}

\usepackage{inconsolata}

\usepackage{graphicx}
\usepackage{amsmath}
\usepackage{amsthm}
\usepackage{booktabs}
\usepackage[ruled,linesnumbered]{algorithm2e}
\usepackage{array}
\usepackage{multirow}
\usepackage{setspace}
\usepackage{subcaption}
\usepackage{bm}

\newcolumntype{P}[1]{>{\centering\arraybackslash}p{#1}}

\title{DMNER: Biomedical Named Entity Recognition by \\ Detection and Matching}

\author{Junyi Bian\textsuperscript{\rm 1 \rm 7}, Rongze Jiang\textsuperscript{\rm 1}, Weiqi Zhai\textsuperscript{\rm 3}, Tianyang Huang\textsuperscript{\rm 3}, Hong Zhou\textsuperscript{\rm 2} \\
{\bf Shanfeng Zhu} \textsuperscript{\rm 3 \rm 4 \rm 5
 \rm 6 \rm 7} \\
\textsuperscript{\rm 1} School of Computer Science, Fudan University, Shanghai 200433, China\\
\textsuperscript{\rm 2} Atypon Systems, LLC, UK\\
\textsuperscript{\rm 3} Institute of Science and Technology for Brain-Inspired Intelligence, Fudan University, China\\
\textsuperscript{\rm 4} Key Laboratory of Computational Neuroscience and Brain-Inspired Intelligence \\ (Fudan University), Ministry of Education, Shanghai 200433, China\\
\textsuperscript{\rm 5} MOE Frontiers Center for Brain Science, Fudan University, Shanghai 200433, China\\
\textsuperscript{\rm 6} Zhangjiang Fudan International Innovation Center, Shanghai 200433, China\\
\textsuperscript{\rm 7} Shanghai Key Lab of Intelligent Information Processing, \\ Fudan University, Shanghai 200433, China\\
\texttt{\{zhusf, 20110240003\}@fudan.edu.cn, hzhou@atypon.com} \\
}

\begin{document}
\maketitle

\begin{abstract}
Biomedical named entity recognition (BNER) serves as the foundation for numerous biomedical text mining tasks. Unlike general NER, BNER require a comprehensive grasp of the domain, and incorporating external knowledge beyond training data poses a significant challenge. In this study, we propose a novel BNER framework called DMNER. By leveraging existing entity representation models SAPBERT, we tackle BNER as a two-step process: entity boundary detection and biomedical entity matching. DMNER exhibits applicability across multiple NER scenarios: 1) In supervised NER, we observe that DMNER effectively rectifies the output of baseline NER models, thereby further enhancing performance. 2) In distantly supervised NER, combining MRC and AutoNER as span boundary detectors enables DMNER to achieve satisfactory results. 3) For training NER by merging multiple datasets, we adopt a framework similar to DS-NER but additionally leverage ChatGPT to obtain high-quality phrases in the training. Through extensive experiments conducted on 10 benchmark datasets, we demonstrate the versatility and effectiveness of DMNER. 
\footnote{The data and code are published in  \url{https://github.com/Eulring/DMNER}.}

\end{abstract}

\section{Introduction}

Biomedical named entity recognition plays a critical role in the field of biomedical natural language processing. 
It serves as the foundation for many biomedical text mining tasks, aiming to effectively handle the vast amount of biomedical text.
Within the biomedical domain, entities are typically defined as a combination of their textual span (name) and their associated category. Common entity categories in the biomedical domain include Disease, Chemical, Species, and others.

Normally, biomedical NER systems have addressed the tasks of detecting entity boundaries and classifying entity types simultaneously. 
For instance, in sequence labeling tasks, the IOB labeling scheme is used to additionally indicate the entity type within the IOB tags.
On the other hand, another approach is commonly seen in nested NER\citep{2003_genia}. It consists of two steps, first identifying the name of the entity (usually in the form of a span), and then using a classifier to determine the category of that span. 
In practice, the ``classifier" is not limited to being trained solely on the current training data, which can also refer to an external model.

\begin{figure}[t]
    \centering
    \includegraphics[width=1\columnwidth]{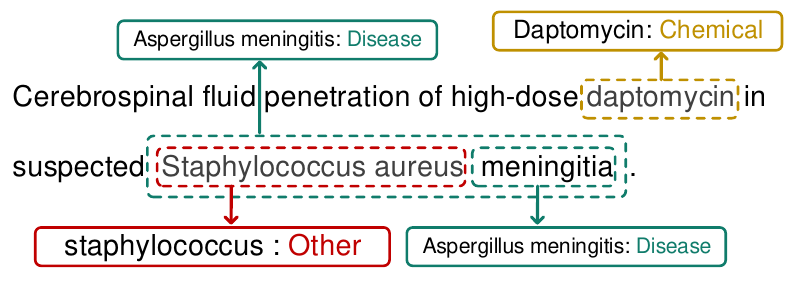}
    \caption{An example from the BC5CDR dataset, where the dashed boxes in the sentence represent entity spans extracted by the EBD module, and the arrows indicate the closest matching entities found by the BEM module.}
    \label{fig:example}
\end{figure}


Recently entity representation models that incorporate contrastive learning and external knowledge bases, like SapBERT \citep{2020_sapbert}, have demonstrated remarkable performance in biomedical concept linking. This task shares a fundamental similarity with the purpose of "classifier" in nested NER. Therefore, we can leverage SapBERT to retrieve the most similar entity to a query entity from a predefined entity dictionary. Under the assumption that similar entities share the same entity type, we can assign the entity type associated with the retrieved entity from the dictionary as the type for the query entity. Therefore, we propose DMNER, which decomposes the BioNER task into two steps: 1) biomedical Entity Boundary Detection (EBD) and 2) Biomedical Entity Matching (BEM).
Figure \ref{fig:example} illustrates an example of DMNER.
The EBD module can be replaced with any entity span extraction model, and the BEM module can be adapted with different dictionaries and entity representation models. This high level of flexibility offered by DMNER allows for its application to various scenarios of supervised NER including distantly supervsied NER (DS-NER) and unified NER.

In DS-NER, where the training data is annotated using an external dictionary, noise is introduced due to the distant labeling process. 
Similarly, in unified NER where multiple datasets are combined for training, each dataset may only be annotated with a subset of possible entity types, leading to many false negative noise labels in the training data. 
Both tasks can essentially be considered as incomplete annotations, which can be addressed by DMNER.
Additionally, given the diverse and potentially overlapping entity types, we employ a nested NER model. During the training of the EBD module, we borrow the idea from AutoNER\cite{2018_autoner} and classify the instances into trusted entities and unknown entities.
The performance of DMNER heavily relies on the recall of the EBD module. To address the low recall problem, for the unified NER task, we leverage external knowledge base UMLS through distant supervision and utilize the annotations from the GPT-3.5 language model to expand the set of unknown entities.

Our contributions can be summarized as follows:

(1) We propose a BNER framework called DMNER, which consists of two steps: Entity Boundary Detection (EBD) and Biomedical Entity Matching (BEM). In the BEM module, we design a dictionary refinement algorithm.

(2) Experimental results on $10$ BNER datasets demonstrate that DMNER further improves the results of supervised NER.

(3) For the DS-NER, we design an EBD model by incorporating previous approaches, where unknown entities are masked during training. Compared to methods without BEM, DMNER achieves a performance improvement of $2.38$ F1-score on the BC5CDR dataset.

(4) In the unified NER task, we employed the same training framework as DS-NER. Additionally, when GPT-3.5 and distant dictionary annotation are introduced to augment the unknown entity set, DMNER performance is significantly improved.

\section{Related Work}


\subsection{Supervised NER}
BiLSTM-CRF \cite{2019_mtm,2018_transfer,2017_habibi} and BERT-based approaches \cite{2020_biobert} have emerged as the paradiam in NER owing to their impressive performance of end-to-end learning. Most of these methods are based on flat NER, which assumes that entity spans do not overlap. In contrast, nested NER \cite{2003_genia} can handle overlapping entities.
Depending on how to formulate the NER task, the NER approaches can be categorized as sequence labeling \cite{2016_ner_task1,2016_ner_task2}, span classification \cite{2018_ner_span1,2019_ner_span2,2021_ner_span3,2017_nest_span}, sequence-to-sequence
\cite{2020_nest_seq2seq},
machine reading comprehension
\cite{2020_mrc}
and contrastive learning
\cite{2022_ner_cl}.
Our approach is similar to span-based methods. While these methods use classifiers for span recognition, we rely on dictionary matching.

\subsection{Distantly-Supervised NER}
Compared to fully supervised NER, DS-NER gets rid of human annotations and uses knowledge bases to annotate the corpus automatically.
Some studies \cite{2020_bond,2019_partial} performed iterative training procedures to mitigate noisy labels in DS-NER.
\citep{2018_autoner, 2019_low, 2019_better} attempted to modify the standard CRF to adapt to label noise scenarios. AutoNER \citep{2018_autoner} introduced an unknown entity dictionary and excluded those entities during training. Building upon AutoNER, Hammer \citep{2020_hamner} and TEBNER \citep{2021_tebner} expanded the dictionary to achieve further enhancements.
In the DS-NER task, our EBD module is also trained based on the idea of masking unknown entities from AutoNER. However, DMNER did not expand the unknown entities dictionary.

\begin{figure*}[t]
    \centering
    \includegraphics[width=1.8\columnwidth]{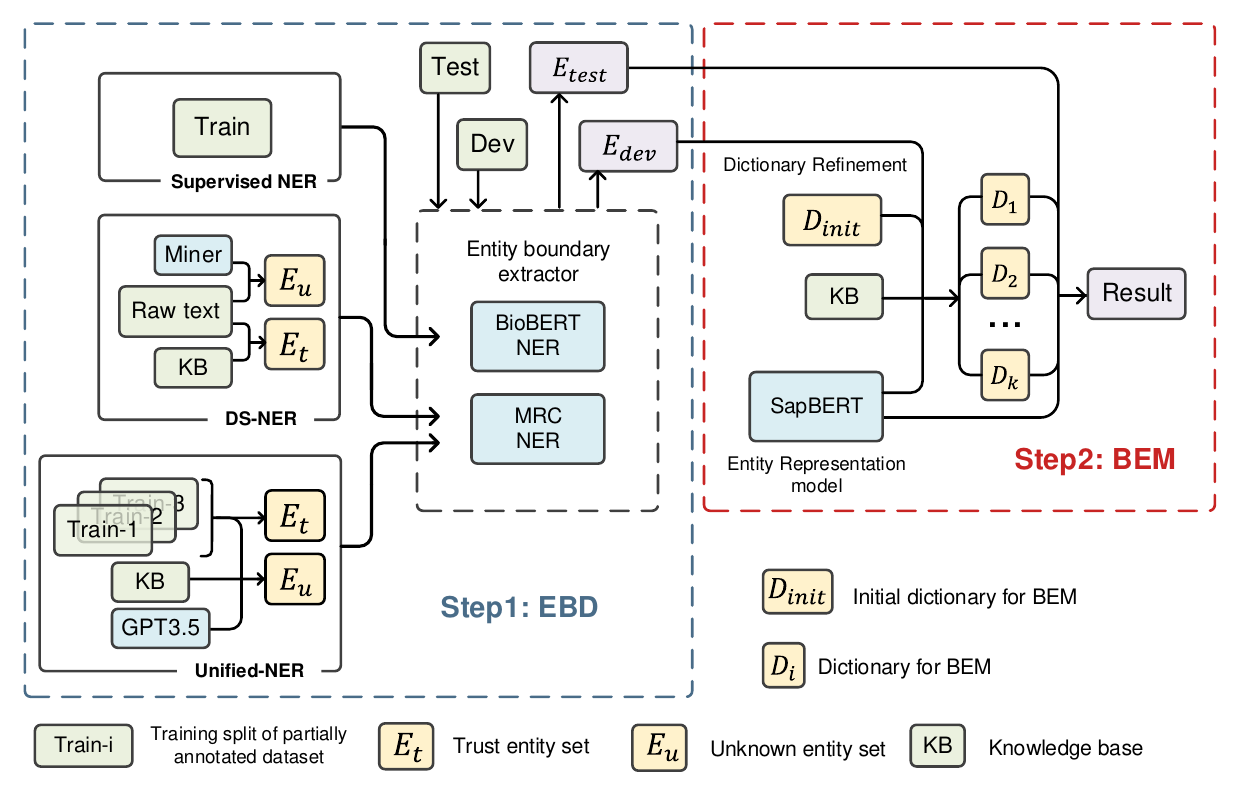}
    \caption{The overall framework of DMNER.}
    \label{fig_framework}
\end{figure*}

\subsection{Unified NER}
Learning from multiple partially annotated datasets can be more broadly conceptualized as learning from imperfect annotations. A major challenge is that a positive instance might be labeled as negative.
A well-explored solution to this problem is proposed by \cite{2008_incomplete}, which only maximizing the likelihood of the gold tag sequence in training. \cite{2018_marginal} applied this idea to build a unified model from multiple partially annotated corpora.
\cite{2019_unified} proposd a crf-based model leverages supervision signals across diverse datasets to learn robust input representations.
The aforementioned methods primarily focus on utilizing supervision signals during training. In contrast, our DMNER relies on external knowledge to filter entities.

\section{Methodology}

\subsection{Task Definition}

\textbf{Tagging Space:} Suppose $D$ is a dataset or an entity set, $T_D$ denote the set of entity types that are tagged in $D$.

\textbf{Named Entity Rcognition:} Given a sentence of words $X=[x_1,x_2,...,x_n]$ from dataset $D$, a NER tagger is required to find all entities. For each entity $e = [x_i,..,x_j]$ from $X$, we denote $<i,j>$ as boundary of the $e$, and $t\in T_D$ as its entity types.

\textbf{Supervised NER:} The tagger is trained on a dataset where each example consists of a sentence $X$ along with the annotations $Y$.

\textbf{Distantly Supervised NER:} The training examples did not provide annotations $Y$,so we use external knowledge base (usually an entity dictionary) to generate pseudo annotations $\mathcal{Y}$ for training.

\textbf{Unified NER:} Given a set of partially annotated datasets ${D_1, ..., D_k}$, those dataset may have different tagging space. Unified NER is to train a single unified NER tagger through the combination of those corpora.
The final trained model is capable of recognizing entity types given by $T_D = T_{D_1} \cup T_{D_2} \cup ... \cup T_{D_k}$.

\subsection{Overall Framework}
DMNER decomposes BNER into two independent steps: Biomedical Entity Boundary Detection (EBD) to extract entity spans and Biomedical Entity Matching (BEM) to determine type of entities extracted from EBD. The models used in these steps can be replaced flexibly, allowing for different approaches.
Figure \ref{fig_framework} illustrates the overall framework of DMNER on three NER tasks.

\subsection{Entity Boundary Detection (EBD)}
The EBD module is essentially equivalent to a traditional NER model, with the distinction that its focus lies in identifying entity boundaries rather than considering entity types.
For datasets containing only one entity type, we employ the flat NER model. However, for NER tasks that involve multiple entity types, with the possibility of overlapping entity mentions, we consistently utilize the nested NER model.

\textbf{Flat NER Tagger:}
We utilize the commonly used BioBERT\cite{2020_biobert} backbone, which has demonstrated strong performance in various NLP tasks in the biomedical domain.

\textbf{Nested NER Tagger:}
Our EBD backbone employs the MRC (Machine Reading Comprehension) \cite{2020_mrc} NER approach.
Before being input into the BioBERT encoding layer, the question ``what is the biomedical entity in this sentence?" and the input sentence X are concatenated, forming the combined string: $[cls]p_1,...,p_m[sep]x_1,...,x_n$. Subsequently, the encoding layer produces the representation matrix $E$, and the MRC NER selects spans by predicting the start and end positions of entities:
\begin{align}
    P_{start} = \text{softmax} (W_{start}\cdot E)
\end{align}
$P_{end} $ is calculated in the same way by replaced trainable matrix $W_{start}$ with $W_{end}$.
Specifically, $P_{start}$ and $P_{end}$ can give the predicted indexes that might be the starting or ending positions, denoted as $\hat{I}_{start}$ and $\hat{I}_{end}$ :

\begin{align}
    \hat{I}_{start} &= \left\{  i | \text{argmax} ( P_{start}^i ) = 1, i=1,...,n  \right\}  \\
    \hat{I}_{end} &= \left\{  i | \text{argmax} ( P_{end}^i ) = 1, i=1,...,n  \right\}
\end{align}
Given any start indexes $i_{st}\in \hat{I}_{start}$ and end indexes $i_{ed}\in \hat{I}_{end}$, a binary classifier is trained to predict the probability that they should be matched:
\begin{align}
    P_{span}^{i_{st},i_{ed}} = \text{sigmoid} (w\cdot \text{concate}(E^{i_{st}}, E^{i_{ed}}))
\end{align}
The entity span $[x_{i_{st}},...,x_{i_{ed}}]$ is then determined based on the value of $P_{span}^{i_{st},i_{ed}}$.


\subsubsection{DS-NER Training}
For the backbone of the distantly supervised approach, we utilize the nested Tagger. In the model training stage, we borrow the idea from AutoNER and introduce two types of entities present in input sentence $X$: trusted entities with corresponding label ($\mathcal{Y}$) and unknown entities with corresponding label ($\hat{\mathcal{Y}}$). For the trusted label ($\mathcal{Y}$), we calculate its loss during training. However, for the unknown entity in the sentence, we ignore its loss contribution.
Trusted entities are obtained through dictionary matching,
while unknown entities are mined using phrase mining tool.
In order to ensure a fair comparison with AutoNER \citep{2018_autoner}, both the set of entities obtained through dictionary-based distant supervision (trusted entities) and the high-quality phrases (unknown entities) are kept consistent with AutoNER.

The loss formulations for the start and end positions are as follows:
\begin{align}
    L_{end} &= \sum_{i=1}^n (1 - \hat{\mathcal{Y}}_{end}^i) CE(P_{end}^i, \mathcal{Y}_{end}^i) \\
    L_{start} &= \sum_{i=1}^n (1 - \hat{\mathcal{Y}}_{start}^i) CE(P_{start}^i, \mathcal{Y}_{start}^i)
\end{align}
Denote $\mathcal{Y}_{span}^{st,ed}$ as the trusted entity labels for whether each start index should be matched with each end index, and the span identification loss is:
\begin{align}
    L_{span} = (1 - \hat{\mathcal{Y}}_{span}^{st,ed}) CE(P_{span}^{st,ed}, \mathcal{Y}_{span}^{st,ed})
\end{align}

The overall training objective is to minimize:
\begin{align}
    L_{DS-NER} = L_{start} + L_{end} + L_{span}
\end{align}

\subsubsection{Unified NER Training}


The framework for unified NER training can also be adapted from the aforementioned DS-NER approach. However, the methods for obtaining two types of entities are different.

For trusted entities, we rely on the partially annotated gold labels provided in the training dataset.
As for unknown entities, they are obtained from two different sources.
One source is distant labeled entities relying on external entity dictionary. The other source is obtained by GPT-3.5. 
In unified NER, due to the wide range of entity types to be recognized, we do not restrict the entity categories during distant annotation and language model annotation. We include any biomedical-related entities as unknown entities, except for those that conflict with trusted entities.
The appendix provides further details on the process of using ChatGPT
for annotating training texts.

\subsection{Biomedical Entity Matching (BEM)}

\textbf{Entity Matching and Filtering}
The previous EBD module extracted potential entities $E=\{ e^i | 1 \leq i \leq N \}$, while the BEM module assigns entity categories to $E$ by matching them against a pre-defined dictionary $\mathcal{D}$:
\begin{align}
    \mathcal{D} = \left \{  <d_{name}^j, d_{type}^j> \big| 1\leq j \leq |\mathcal{D}|    \right \}
\end{align}
The dictionary $\mathcal{D}$ consists of carefully selected entities, with each entity comprising its name $d_{name}$ and corresponding entity category $d_{type}$.
In the BEM phrase, each entity $e^i$ is matched against the entity names in the dictionary $\mathcal{D}$ to select one with the highest similarity score. The entity category corresponding to the selected entity is then assigned as the entity type for $e^i$. This process can be formulated as:
\begin{align}
    BEM(e^i, \mathcal{D}) = d_{type}^{ \mathop{\arg\max}\limits_{j} (sim(e^i, d^j_{name}))}
\end{align}
DMNER utilizes SapBERT \cite{2020_sapbert} to encode the textual representations of entitie $e^i$ and $d_{name}^j$. We then compute the cosine similarity between the encoded vectors as the similarity score between the two entities.
Note if the predicted entity type $BEM(e^i, \mathcal{D})$ is not present in the tagging space $T_{E}$, we will discard this prediction. This approach aims to improve the precision of the pipelined result.

\textbf{Dictionary Construction} The dictionary plays a crucial role in the BEM module, providing essential entity information. The primary source of entities for the dictionary is the training data of the EBD model. However, these entities are often limited to the labeled tagging space $T$. When the EBD module extracts entities of other types, they cannot be filtered, resulting in a restricted number of entities. To address this issue, we can enhance the dictionary by incorporating entities from external knowledge bases such as UMLS and CTD.
To determine which entities to include, we leverage the EBD outputs of the validation split $E_{dev}$ along with the true entity types associated with the $E_{dev}$. Specifically, we add entities to the dictionary if their selection improves the BEM performance on the development set. To achieve this, we have devised a greedy search-based dictionary refinement algorithm, enabling the expansion of the initial dictionary.
\begin{align} \label{eq:s3-123}
    \mathcal{D} = \text{Refine} (\mathcal{D}_{init}, KB, E_{dev}, Y_{dev})
\end{align}

The algorithm undergoes multiple iterations. In each iteration, it begins by selecting a fixed number of entities $E_b$ from the knowledge base $KB$. For each entity $e_b$ in $E_b$, the algorithm identifies all units in  $E_{dev}$ whose most similar entity is $e_b$. These units are dentoed as $E_{e_b}$. Next, it calculates the count of entities in $E_{e_b}$ that share the same entity type as $e_b$ ($\text{cnt}_p$), as well as the count of entities with different types ($\text{cnt}_n$). If $\text{cnt}_p$ exceeds $\text{cnt}_n$, the entity is added to the dictionary ($\mathcal{D}$). When entities are added to $\mathcal{D}$, some existing entities may be removed. 
For each entity $e_d$ in $\mathcal{D}$, the algorithm performs a similar process to obtain $E{e_d}$ and calculates the difference in matching. If $\text{cnt}_n$ exceeds $\text{cnt}_p$ by a threshold value ($t$), $e_d$ is removed from $\mathcal{D}$. The threshold is set because removing entities may result in new matches for $E{e_d}$ in the dictionary, but these matches cannot guarantee correctness.
For more detailed information, please refer to Algorithm \ref{alg:algorithm}. In eq \ref{eq:s3-123}, $\mathcal{D}_{init}$ represents the initial dictionary. The selection of $\mathcal{D}_{init}$ will be discussed in Section 4.2.

\begin{algorithm}[tb]
    \caption{Dictionary Refinement}
    \label{alg:algorithm}
    \KwIn{$E_{dev}$,$Y_{dev}$,${KB}$,$\mathcal{D}_{init}$}
    \KwOut{$\mathcal{D}$}
    Let $\mathcal{D} = \mathcal{D}_{init}$ \\
    \For{$i \leftarrow 1$ to $iter$}{
        sample a batch of entities $E_b$ from ${KB}$ \\
        \ForEach{$e_b$ in $E_b$}{
            Get $E_{e_b}$ from $E_{dev}$ based on $e_b$ \\
            $\text{cnt}_{p}=\#$ same type entities in $E_{e_b}$ \\
            $\text{cnt}_{n}=\#$ diff. type entities in $E_{e_b}$ \\
            \If{$\text{cnt}_{p} > \text{cnt}_{n}$}{
                add $e_b$ into $\mathcal{D}$
            }
        }
        \ForEach{$e_d$ in $\mathcal{D}$}{
            Get $E_{e_d}$ from $E_{dev}$ based on $e_d$ \\
            $\text{cnt}_{p}=\#$ same type entities in $E_{e_d}$ \\
            $\text{cnt}_{n}=\#$ diff. type entities in $E_{e_d}$ \\
            \If{$\text{cnt}_{n} > \text{cnt}_{p} + \text{t}$}{
                remove $e_d$ from $\mathcal{D}$
            }
        }

    }
\end{algorithm}

\textbf{Reuslt Ensembling}
To further enhance the model's performance, we construct multiple dictionaries for each test dataset. The BEM module predicts a result using each dictionary, and these results are aggregated through voting to determine the final prediction. Since the $KB$ is extensive, we access only a subset of it during dictionary construction. Therefore, we shuffle the $KB$ to obtain multiple distinct dictionaries.




\begin{table*}[htbp] \small
  \centering
  \begin{tabular}{P{26mm}|P{8mm}|P{10mm}P{12mm}|P{10mm}P{10mm}|P{10mm}P{12mm}|P{12mm}p{10mm}}
    \toprule
    \multirow{2}{*}{Model} & \multirow{2}{*}{Metric} & \multicolumn{2}{c|}{Disease} & \multicolumn{2}{c|}{Gene} & \multicolumn{2}{c|}{Chemical} & \multicolumn{2}{c}{Species} \\
                &  &  NCBI  & BC5CDR  & BC2GM  &  JNLPBA & BC4CH & BC5CDR  &  Linnaeus & S800  \\
    \midrule
    \multirow{3}{*}{BioBERT NER}
                &P& 86.61 & 82.05  & 81.07 & 70.45  & 90.16 & 91.16  & 90.89 & 67.48 \\
                &R& 89.68 & 85.69  & 82.83 & 83.05  & 88.95 & 92.83  & 85.62 & 75.48 \\
                &F& \underline{88.12} & 83.83  & 81.94 & 76.23  & \underline{89.55} & \underline{91.98}  & 88.17 & 71.26 \\ \midrule
    \multirow{3}{*}{$\text{DMNER}_S$}
                &P& 87.24 & 85.18  & 81.83 & 70.80  & 91.48 & 93.78  & 91.40 & 68.43 \\
                &R& 88.82 & 83.04  & 82.47 & 82.89  & 87.62 & 90.12  & 85.51 & 75.44 \\
                &F& 88.02 & \underline{84.10}  & \underline{82.15} & \underline{76.37}  & 89.50 & 91.92  & \underline{88.35} & \textbf{71.76} \\ \midrule
    \multirow{3}{*}{$\text{DMNER}_E$}
            &P& 87.30 & 85.44  & 81.78 & 70.82  & 91.50 & 93.75  & 91.43 & 68.27 \\
            &R& 89.58 & 83.54  & 82.61 & 83.03  & 87.81 & 90.54  & 85.62 & 75.48 \\
            &F& \textbf{88.43} & \textbf{84.48}  & \textbf{82.19} & \textbf{76.44}  & \textbf{89.61} & \textbf{92.12}  & \textbf{88.43} & \underline{71.70} \\
    \bottomrule
  \end{tabular}
  \caption{Precision (P), Recall (R) and F1 (F) scores on supervised NER dataset are reported.
  The best F1 scores are in bold, and the second best scores are underlined.} \label{tab:supervised}
\end{table*}

\begin{table}[htbp] \small
  \centering
  \begin{tabular}{l|p{11mm}p{11mm}p{11mm}}\toprule
        method                &  Pre  &  Rec  &  F1    \\ \midrule
        Dictionary Match      & 93.93 & 58.35 & 71.98   \\
        AutoNER              & 83.08 & 82.16 & 82.70   \\
        MRC-mask              & 79.48 & 87.33 & 83.22   \\  
        $\text{DMNER}_S\text{-}\mathcal{D}{init}$    & 85.15 & 85.61 & 85.38   \\ 
        $\text{DMNER}_E$           & 87.96 & 84.20 & 86.04   \\
        $\text{DMNER}_S$           & 87.79 & 83.52 & 85.60   \\ \midrule
        \textit{DMNER-EBD}  & \textit{85.37} & \textit{87.14} & \textit{86.24}   \\
        \bottomrule
  \end{tabular}
   \caption{DS-NER results on BC5CDR dataset.} \label{tab:dsner}
\end{table}

\begin{table}[htbp] \small
  \centering
  \begin{tabular}{l|p{9mm}p{9mm}p{9mm}}\toprule
        method                &  Pre  &  Rec  &  F1    \\ \midrule
        MTM vote              & 64.4  & 62.8  & 63.6   \\
        Unified  CRF          & 84.1  & 65.7  & 73.8   \\
        $\text{DMNER}_E\text{-}\mathcal{D}{init}$&  81.8 & 67.3  &  73.8   \\
        $\text{DMNER}_S\text{-}\mathcal{D}{init}$&  81.6 & 66.6  &  73.3   \\
        $\text{DMNER}_E$              &  85.8 & 65.6  &  74.3   \\
        $\text{DMNER}_S$              &  85.6 & 64.5  &  73.6   \\ \midrule
        \textit{DMNER-EBD}             & \textit{73.2}  & \textit{70.7}  & \textit{72.0}   \\
        \bottomrule
  \end{tabular}
   \caption{Unified NER Results on BC5CDR. } \label{tab:unify}
\end{table}

\section{Experiments}

\subsection{Datasets}

The \textbf{Supervised NER} datasets, following BioBERT \cite{2020_biobert}, encompass various datasets such as BC5CDR-chem \cite{2016_bc5cdr} and BC4CHEMD \cite{2015_bc4chemdner} for chemical entities, s800 \cite{2013_s800}  and linnaeus \cite{2010_linnaeus} for species entities, JNLPBA
\cite{2004_jnlpba}
and BC2GM
\cite{2008_bc2gm}
for Gene/Protein entities, and BC5CDR-disease \cite{2016_bc5cdr} and NCBI \cite{2014_ncbi} for disease entities.

\textbf{DS-NER} was evaluated on the BC5CDR dataset, where trusted entities were obtained from MeSH
\footnote{\url{www.nlm.nih.gov/mesh/download_mesh.html}}
and CTD
\footnote{\url{http://ctdbase.org/downloads/}}
. Unknown entities were mined using AutoPhrase \cite{2018_autophrase}.

In \textbf{Unified NER}, the EBD model combined five datasets, namely BC4CHEMD (Chemical), JNLPBA (Gene), BC2GM (Gene), NCBI (Disease), and Linnaeus (Species), for training purposes.
Evaluation was conducted on a new datasets, BC5CDR (Chemical and Disease).

\subsection{Experiments Setting}
In this paper, the knowledge base (KB) has two purposes. The first one is to build the dictionary for the BEM module, and the second one is to expand the trusted entities used in the training of unified NER.
The knowledge bases sourced from UMLS
\footnote{\url{https://download.nlm.nih.gov/umls/kss/2023AA/umls-2023AA-full.zip}}
and CTD. Further details about them can be found in the Appendix.

During dictionary refinement, consistent hyperparameters are utilized across all test datasets. In the algorithm, the values are set as follows: $t=2$, $iter=20$, and each iteration samples $4096$ entities.
The experiments of parameter selection are presented on Appendix.

In supervised NER, $\mathcal{D}_{init}$ was derived from the training data. In DS-NER and Unified NER, $\mathcal{D}_{init}$ was sampled from trusted entities.
The flat NER tagger of the EBD module was trained using the same settings as BioBERT\cite{2020_biobert}, whereas the nested NER tagger was trained using the same settings as MRC\cite{2020_mrc}.

\begin{table*}[htbp] \small
  \centering
  \begin{tabular}{P{32mm}|P{15mm}|P{8mm}|P{8mm}P{8mm}P{8mm}|P{8mm}|P{8mm}P{8mm}P{8mm}}
    \toprule
    \multirow{2}{*}{Dataset} & Model:  & \multicolumn{4}{c|}{Random} & \multicolumn{4}{c}{Standard} \\
                             & Metric: &  dsize  &  Pre  &  Rec  &  F1  &  dsize  &  Pre  &  Rec  &  F1  \\
    \midrule
    \multirow{2}{*}{Supervised NER: NCBI}
                            & before DR & 1707 & 87.81  & 64.45 & 74.34  & 1707 & 86.61  & 100.0 & 92.83 \\
                            & after DR  & 2272 & 87.90  & 88.61 & 88.26  & 1843 & 87.40  & 99.88 & 93.22 \\ \midrule
    \multirow{2}{*}{DS-NER: BC5CDR}
                            & before DR & 2482 & 91.48  & 56.69 & 70.00  & 2482 & 89.71  & 90.31 & 90.01 \\
                            & after DR  & 4768 & 91.69  & 89.31 & 90.48  & 3572 & 88.61  & 95.37 & 91.87 \\ \midrule
    \multirow{2}{*}{Unified NER: BC5CDR}
                            & before DR & 4600 & 85.35  & 59.73 & 70.28  & 4600 & 84.00  & 93.93 & 88.69 \\
                            & after DR  & 7425 & 88.35  & 86.04 & 87.18  & 6721 & 88.11  & 92.30 & 90.16 \\
    \bottomrule
  \end{tabular}
  \caption{Impact of different $\mathcal{D}_{init}$ on the BEM results and the influence of dictionary refinement (DR) on the BEM results were evaluated across three NER tasks. "dsize" represents the size of the dictionary used for BEM. "Random" means $\mathcal{D}_{init}$ is randomly selected from the knowledge base, while "standard" refers to the default initialization of dictionary.} \label{tab:dr}
\end{table*}

\subsection{Baselines}
\textbf{Supervised NER:}
We adopt the \textbf{BioBERT NER} model as our baseline. Considering that different dictionaries can impact the model's performance, we employ the dictionary refinement algorithm to generate $k$ different dictionaries and evaluate their respective results. By averaging the scores obtained from these evaluations, we obtain $\text{\textbf{DMNER}}_{\bf{S}}$.
On the other hand, $\text{\textbf{DMNER}}_{\bf{E}}$ integrates the results obtained from the aforementioned $k$ dictionaries to achieve an ensemble-based approach.

For a fair comparison, the EBD module in DMNER is identical to the BioBERT NER. The distinction between  DMNER and BioBERT NER lies in the addition of the BEM module in DMNER.
In the case of BioBERT, the model was trained by combining the training and development data. However, in our study, we retrained the model exclusively using the training split of the data.

\textbf{DS-NER:}
The EBD module in DMNER is built upon MRC NER and adheres to the training strategy outlined in section 3.3.1. It is important to note that in the strict settings of DS-NER and unified NER, the dev split lacks gold labels, rendering dictionary refinement infeasible. To overcome this limitation, we introduced the
 $\text{\textbf{DMNER-}}\bm{\mathcal{D}_{init}}$ method for comparison. This method directly employs $\mathcal{D}_{init}$ as $\mathcal{D}$ for BEM, eliminating the necessity for $Y_{dev}$.
Furthermore, owing to the limited number of trusted entities in the DS-NER dataset, $\mathcal{D}_{init}$ is distinct and includes all trusted entities. As a result, there is no ensemble baseline $\text{DMNER}_E\text{-}\mathcal{D}_{init}$.
The original AutoNER model employs all raw texts for training, so, we rely on the results from \cite{2020_hamner}, who retrained AutoNER using only the training split, to ensure a fair comparison.
The primary distinction between the \textbf{MRC-mask} and DMNER is that MRC-mask does not incorporate the BEM module. Instead, it directly recognizes entity boundaries and types in a single stage.

\textbf{Unified NER:} \textbf{Unified CRF} are method in the paper \cite{2019_unified}. \textbf{MTM Vote}\cite{2019_mtm} trains an NER tagger for each dataset using multi-task learning and then votes on the results.
In both DS-NER and unified NER, we also documented the results of \textbf{DMNER-EBD}, representing the performance of the first stage in DMNER. The first stage EBD model solely focuses on identifying entity boundaries and is included in the table for comparison purposes with DMNER. In supervised NER, the EBD results of $\text{DMNER}$ are essentially identical to those of BioBERT NER.

\subsection{Overall Results}

The results of DMNER on supervised NER are presented in Table \ref{tab:supervised}.
$\text{DMNER}_{S}$ refers to the results obtained using a single dictionary, whereas
$\text{DMNER}_{E}$ refers to the results obtained using multiple dictionaries.
Across all datasets, DMNER emerges as the best performing model. Specifically, $\text{DMNER}_{E}$  exhibits the highest improvement on BC5CDR-disease, with a notable increase of 0.63 in F1-score. Moreover, $\text{DMNER}_{S}$ surpasses BioBERT NER on most datasets. However, there is an exception with the S800 dataset, where $\text{DMNER}_{E}$ performs slightly lower than $\text{DMNER}_{S}$. This discrepancy could potentially be attributed to the performance of a specific dictionary, which impacts the overall integration of results.
The key distinction between $\text{DMNER}$ and BioBERT lies in the inclusion of the BEM module within DMNER. This observation implies that the BEM module plays a crucial role in refining and rectifying entity predictions in supervised NER, leading to improved performance.


Table \ref{tab:dsner} shows the results of DS-NER.
The MRC-mask model achieved an impressive F1 score that surpasses AutoNER by 0.52, underscoring the effectiveness of the MRC framework.
Notably, the introduction of the BEM module in $\text{DMNER}_S\text{-}\mathcal{D}{init}$
led to a substantial improvement of 2.16 compared to MRC-mask, demonstrating the remarkable effectiveness and impact of DMNER in DS-NER.

Table \ref{tab:unify} shows the results of unified NER. Remarkably, the performance of $\text{DMNER}_S\text{-}\mathcal{D}{init}$ is comparable to that of the previous state-of-the-art unified CRF model. 
However, our method is more flexible as it allows us to recognize entities that are beyond the tagging space of the training data by setting entities in the BEM dictionary.
This clearly demonstrates the feasibility and effectiveness of our approach in the unified NER task.

In all the comparisons between DMNER and DMNER-EBD, it can be observed that DMNER achieves slightly lower recall compared to the EBD model. This is due to the inherent limitation of DMNER's recall being dependent on EBD's recall. However, DMNER generally demonstrates improved precision.




\subsection{Impact of dictionary construction}

Table \ref{tab:dr} presents the impact of dictionaries on the BEM module, focusing on the evaluation metrics specific to the BEM stage rather than the overall results of DMNER.
 The ``random" method randomly selects the same number of entities from the $KB$ as the standard method and undergoes dictionary refinement (DR). Based on the findings from the table, the following observations can be made:
1) Applying DR leads to improved results for the BEM stage, especially for randomly initialized dictionaries.
2) The initialization of the dictionary significantly affects the performance of the BEM module.
3) Due to the poor performance of $\mathcal{D}_{init}$, DR extracts more entities to compensate for the weaknesses in recall.

\begin{figure}[htbp]
    \centering
    \includegraphics[width=1\columnwidth]{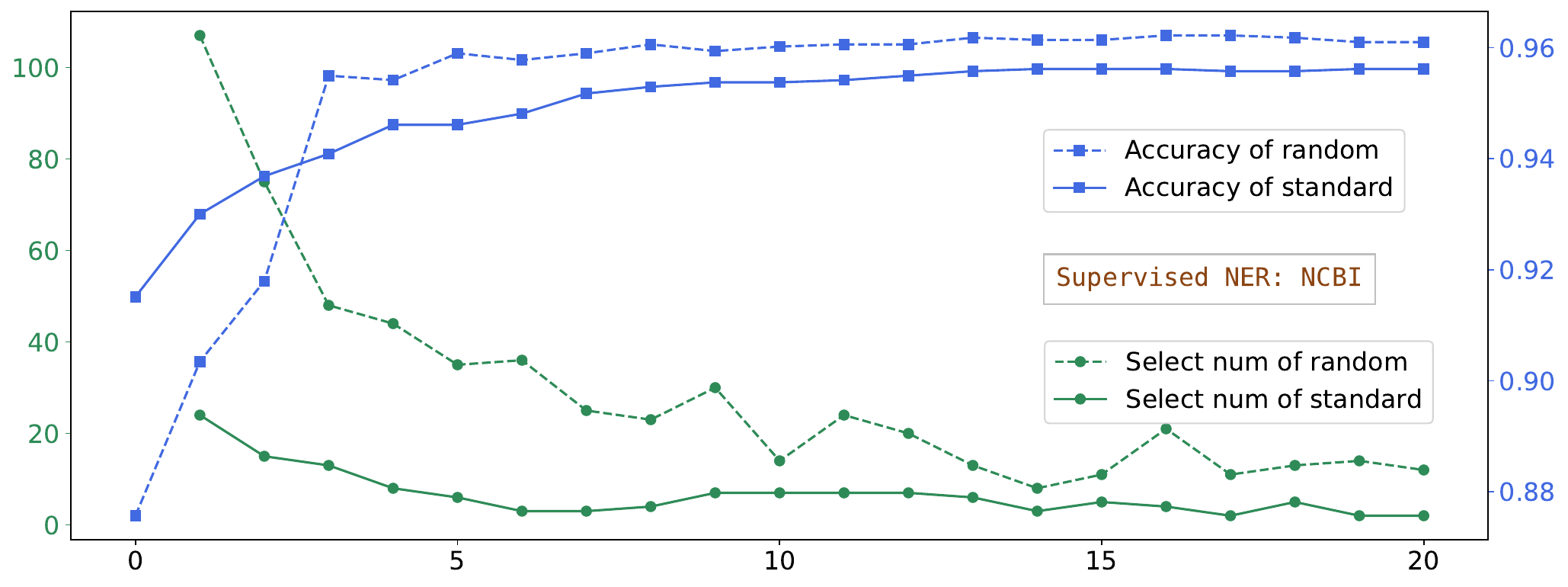}
    \caption{The accuracy on $E_{dev}$ and the number of samples per iteration for the dictionary refinement algorithm using different dictionary initializations are shown for each iteration. The green line represents the progressively increasing number of entities during the iterations, while the blue line represents the accuracy. Solid lines indicate initialization with a specific entity set, and dashed lines represent random initialization.}
    \label{fig:aba2}
\end{figure}

Figure \ref{fig:aba2} illustrates the accuracy changes of $\mathcal{D}$ on $E_{dev}$
throughout the iterations of the DR algorithm, along with the number of entities selected to be added to  $\mathcal{D}$  from the candidate pool at each iteration
When comparing the random method to the standard method, it is observed that the random method initially displays considerably lower accuracy but selects a larger number of entities in each iteration. Towards the end, the accuracy may even slightly improve. However, as indicated in Table \ref{tab:dr}, the final results are worse, suggesting overfitting.

\subsection{Impact of Trusted entity in unifed NER}

In unified NER training, there are two sources of unknown entities. One is obtained through distant supervision using the knowledge base, while the other is annotated using GPT-3.5. In Table 6, the comparison between DMNER and other variations is presented. \textbf{DMNER w/o GPT} refers to DMNER's EBD module trained without using unknown entities generated by GPT-3.5. Similarly, \textbf{DMNER w/o KB} indicates that remote labels from the knowledge base are not used to form unknown entities. In \textbf{DMNER w/o unk.}, the unknown entity category is completely empty.
The experimental results reveal that excluding unknown entities leads to a decrease in model recall by 34.11 due to the presence of numerous false positives in the training data. However, the model's performance improves significantly when GPT labels or remote labels from the knowledge base are included in the unknown entity category. Among these variations, DMNER w/o GPT achieves a slightly lower result by 0.63 compared to DMNER w/o KB, suggesting that GPT-labeled entities exhibit higher quality compared to distantly labeled entities.


\begin{table}[tbp] \small
  \centering
  \begin{tabular}{l|p{11mm}p{11mm}p{11mm}}\toprule
        method                &  Pre  &  Rec  &  F1    \\ \midrule
        $\text{DMNER}_E$                 & 85.83 & 65.55 & 74.33   \\
        \quad w/o GPT      & 85.60 & 54.74 & 66.78   \\
        \quad w/o KB          & 89.53 & 54.06 & 67.41   \\
        \quad w/o unk     & 89.79 & 25.93 & 40.24   \\
        \bottomrule
  \end{tabular}
   \caption{Abalation study of different sources of unknown entities in the unified NER model training.} \label{tab:aba1}
   \vspace{-3mm}
\end{table}



\section{Conclusion}

We propose a novel solution for Biomedical Named Entity Recognition called DMNER. DMNER consists of two modules, BEM (Biomedical Entity Matching) and EBD (Entity Boundary Detection), which can be flexibly adapted to different NER scenarios, such as supervised NER, DS-NER, and Unified NER.
By directly configuring the dictionary in the BEM module, DMNER can recognize various types of entities. 
DMNER is a relatively new framework and offers significant potential for further exploration. 

\bibliography{main}
\bibliographystyle{acl_natbib}
\newpage

\appendix

\section{Appendix}

\subsection{Dataset Details}
The detailed statistics of BioNER datasets are shown in Table \ref{tab:dataset}.
The dataset used in Supervised NER is consistent with the one utilized in the paper \cite{2020_biobert}. 
Additionally, the BC5CDR dataset used in DS-NER is consistent with the dataset used in AutoNER\cite{2018_autoner}. Similarly, the BC5CDR dataset used in unified NER is also consistent with the dataset used in Unified CRF\cite{2019_unified}.
\begin{table}[htbp] \small
  \centering
  \begin{tabular}{p{17mm}|p{6mm}|p{8mm}|p{8mm}|p{8mm}}
    \toprule
    Dataset & Type & Train & Dev & Test\\
    \midrule
    \multicolumn{5}{c}{Supervised NER}\\
    \midrule
    BC5CDR-C     & C & 4582  & 4602  & 4812\\ 
    BC4CHEMD     & C & 30884 & 30841 & 26561\\
    s800         & S & 5743  & 831   & 1630\\
    linnaeus     & S & 12004 & 4086  & 7181\\
    JNLPBA       & G & 14731 & 3876  & 3873\\
    BC2GM        & G & 12632 & 2531  & 5065\\
    BC5CDR-D     & D & 4582  & 4602  & 4812\\
    NCBI         & D & 5432  & 923   & 942\\ 
    \midrule
    \multicolumn{5}{c}{Distantly supervised NER}\\ 
    \midrule
    BC5CDR       & C D & 4559  & 4580  & 4796 \\ 
    \midrule
    \multicolumn{5}{c}{Unified NER}\\ 
    \midrule
    BC5CDR       & C D & 4559  & 4580  & 4796 \\ 
    \bottomrule
  \end{tabular}
  \caption{The statistics of the BNER datasets. BC5CDR-C refers to the BC5CDR chemical dataset, while BC5CDR-D represents the BC5CDR disease dataset. In the entity categories, C denotes Chemical, D denotes Disease, G denotes Gene/Protein, and S denotes Species. }
 \label{tab:dataset}
\end{table}

\subsection{ChatGPT Labeling}

In unified NER, one approach to improve the recall of the EBD module is to expand the set of unknown entities during training.
Larger language model can be used to achieve this.
We utilize the ChatGPT3.5 API to annotate the training in the following three steps:

\begin{itemize}
    \item (1) Constructing a prompt: \textit{Please extract all biomedical entities as much as possible while ignoring other types of entities. Each entity a line and start with '-', you do not need to point out entity type.}
    \item  (2) Feeding the constructed prompt along with sentence to GPT-3.5-Turbo model to obtain the generated GPT answer;
    \item  (3) Parsing the GPT answer into entity spans.
\end{itemize}
In addition, due to the variability in the output of ChatGPT, we performed three annotations for each example and used a voting scheme to determine the final result. This approach helped to mitigate the potential instability in the ChatGPT outputs and ensure more reliable annotations.

Table \ref{tab:appendix2} presents an example of extracting biomedical entities from training examples using the ChatGPT API.

\begin{table}[htbp] \small
  \centering
  \begin{tabular}{p{70mm}}
    \toprule
    Prompt\\\midrule
    Please extract all biomedical entities as much as possible while ignoring other types of entities. Each entity a line and start with '-', do not need to point out entity type.\\\midrule
    Sentence\\\midrule
    Although all of the currently available H2-receptor antagonists have shown the propensity to cause delirium, only two previously reported cases have been associated with famotidine.\\\midrule
    GPT Answer\\\midrule
    - H2-receptor antagonists\\
    - delirium\\
    - famotidine\\\midrule
    Entity Spans\\\midrule
    GPT span: [6:7], [14:14], [25:25]\\
    Gold span: [14:14], [25:25]\\
    \bottomrule
  \end{tabular}
  \caption{An exmaple of ChatGPT labeling on BC5CDR.}
 \label{tab:appendix2}
\end{table}

\begin{table*}[htbp] \small
  \centering
  \begin{tabular}{p{25mm}|p{13mm}|p{13mm}|p{13mm}|p{13mm}|p{13mm}|p{13mm}|p{13mm}}
    \toprule
    Knowledge Base & Anatomy & Pathway & Disease & Gene & Chemical & Species & Other \\
    \midrule
    CDT  & 1844 & 2567 & 13 270 & 589 822 & 175 910 & - & -\\ \midrule
    UMLS & 343 712 & - & 616 025  & 344 979 & 343 011  & 1355 284 & 5453 977 \\
    \bottomrule
  \end{tabular}
  \caption{The statistic categories and corresponding entities from the CTD and UMLS knowledge bases.} \label{tab:appendix1}
\end{table*}

\begin{figure}[h]
    \centering
    \begin{subfigure}[b]{0.23\textwidth}
        \includegraphics[width=\linewidth]{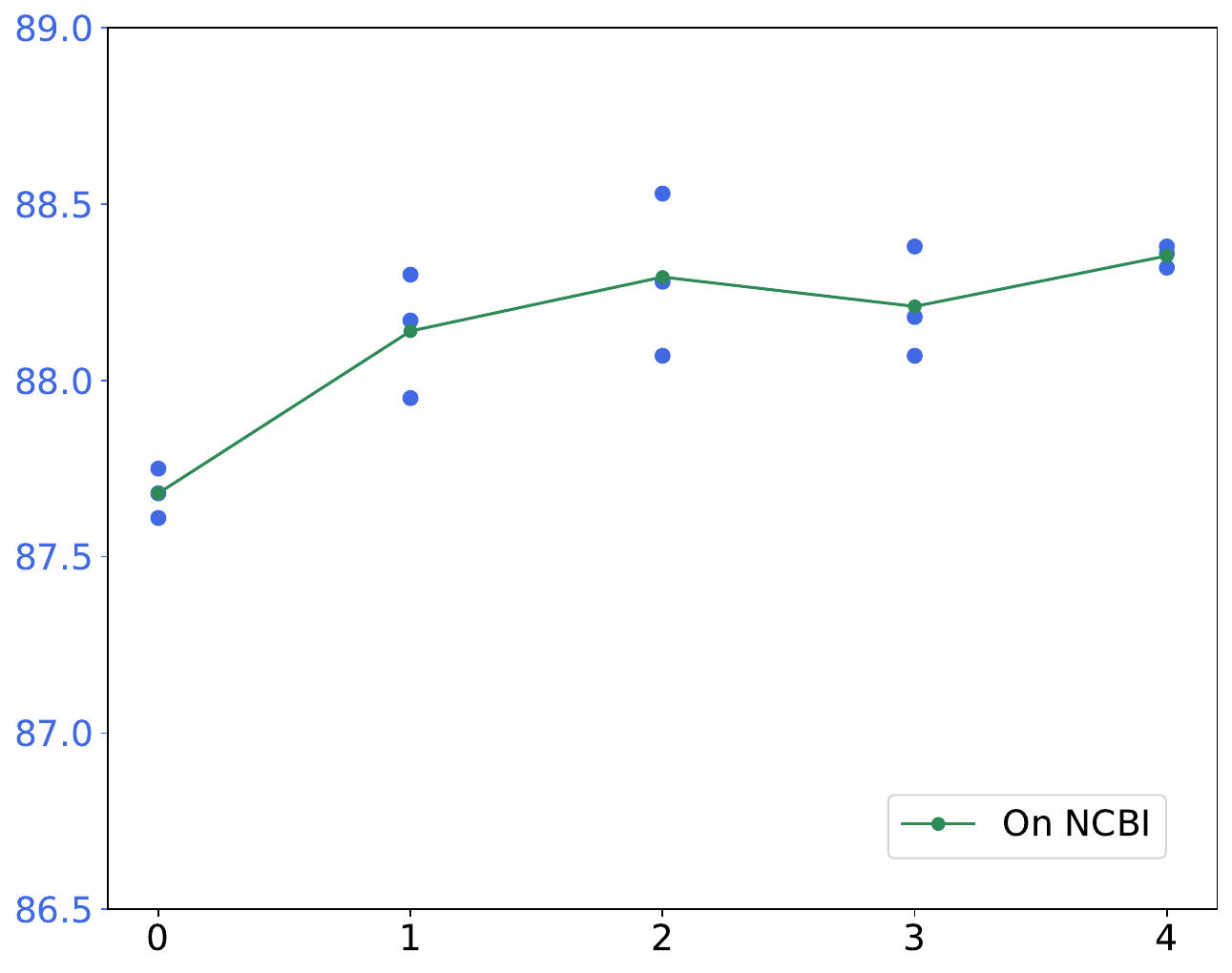}
        \caption{}
        \label{fig:aba3-1}
    \end{subfigure}
    \hfill
    \begin{subfigure}[b]{0.23\textwidth}
        \includegraphics[width=\linewidth]{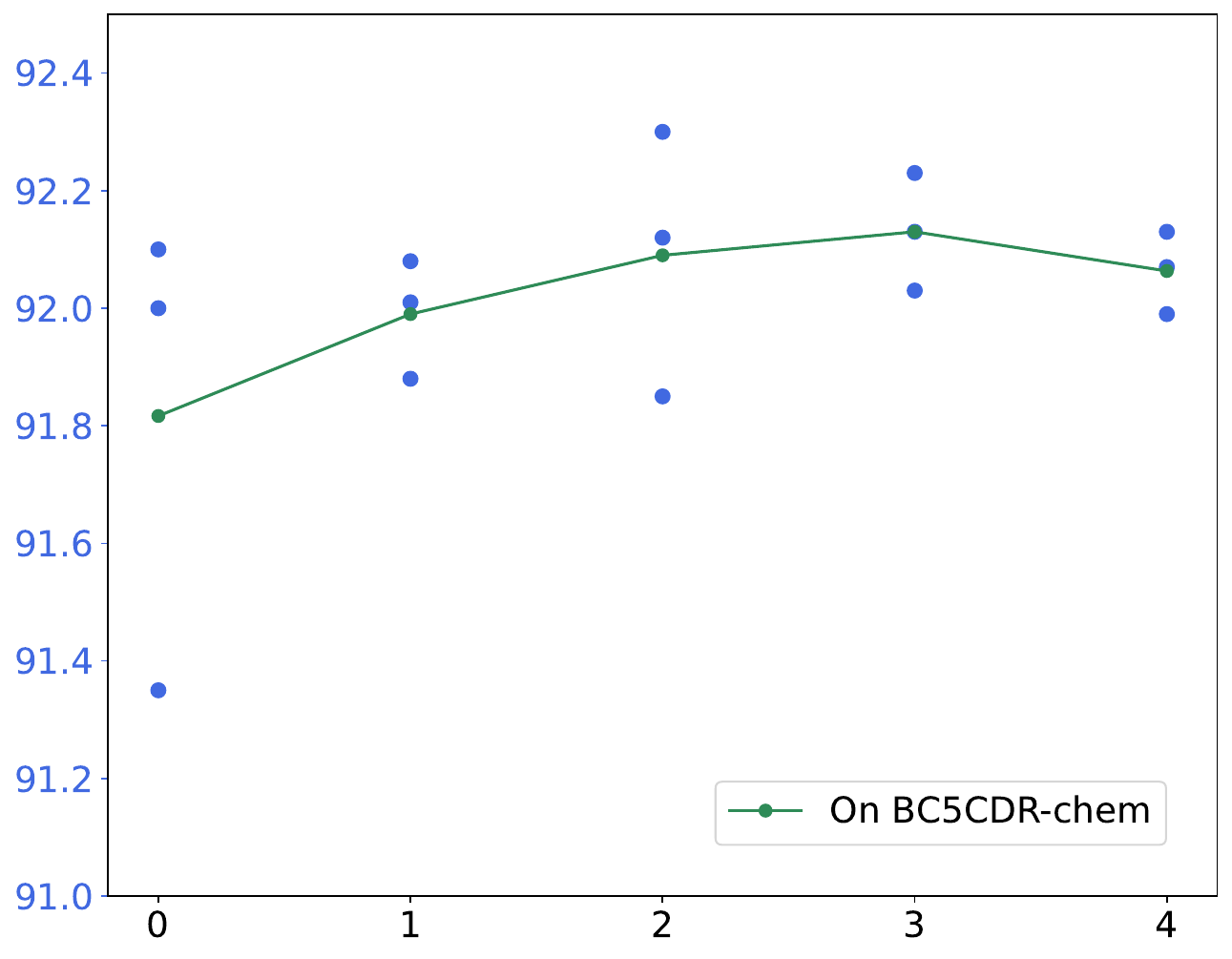}
        \caption{}
        \label{fig:aba3-2}
    \end{subfigure}
    \caption{The influence of different values of $t$ in the dictionary refinement algorithm on the model's performance across the three NER tasks is shown in the plot. The x-axis represents the values of $t$, while the y-axis represents the score of the DMNER. (a) and (b) represent the experiments conducted on the NCBI and BC5CDR-chem datasets, respectively.}
    \label{fig:aba3}
\end{figure}

\begin{figure}[ht]
    \centering
    \begin{subfigure}[b]{0.23\textwidth}
        \includegraphics[width=\linewidth]{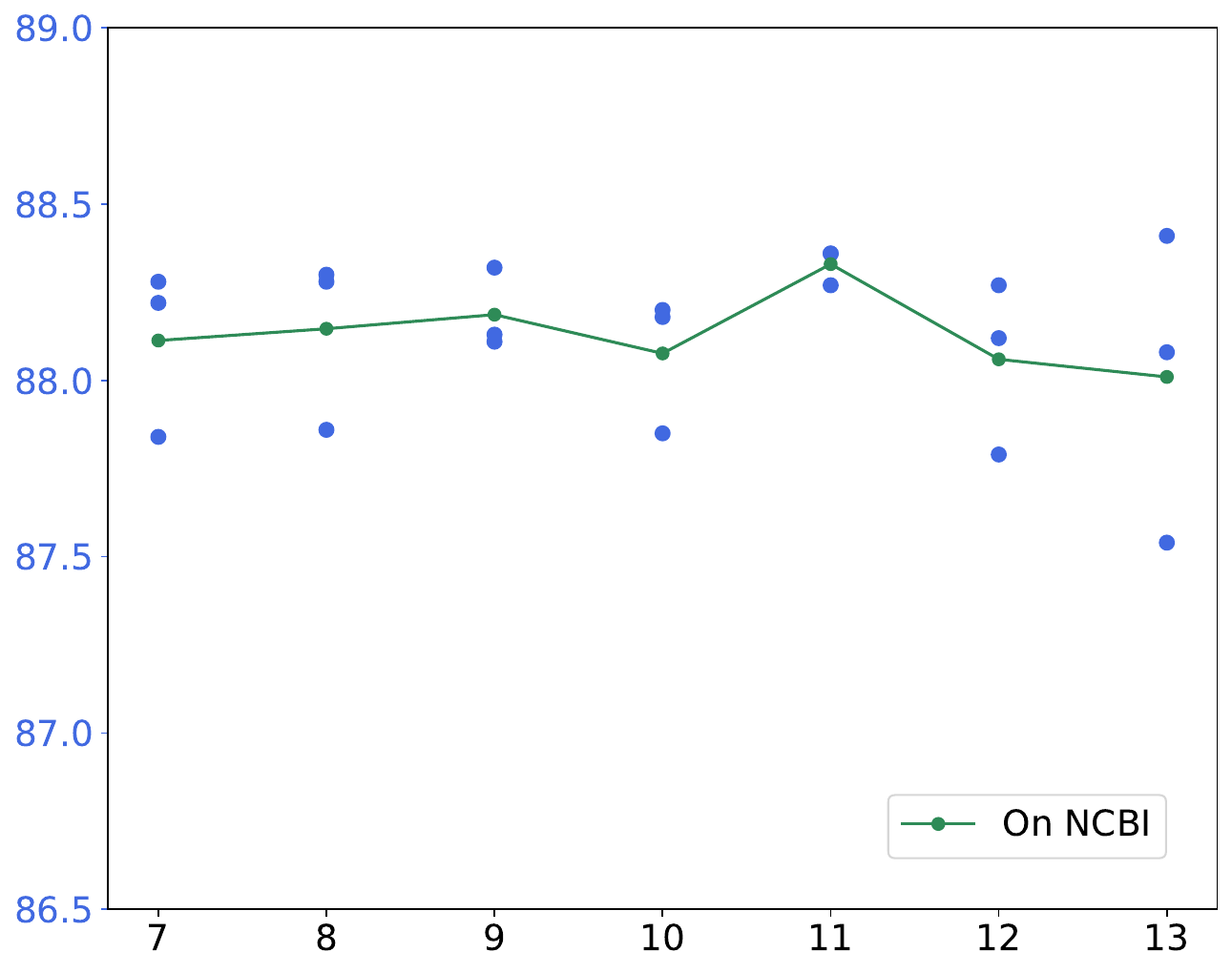}
        \caption{}
    \end{subfigure}
    \hfill
    \begin{subfigure}[b]{0.23\textwidth}
        \includegraphics[width=\linewidth]{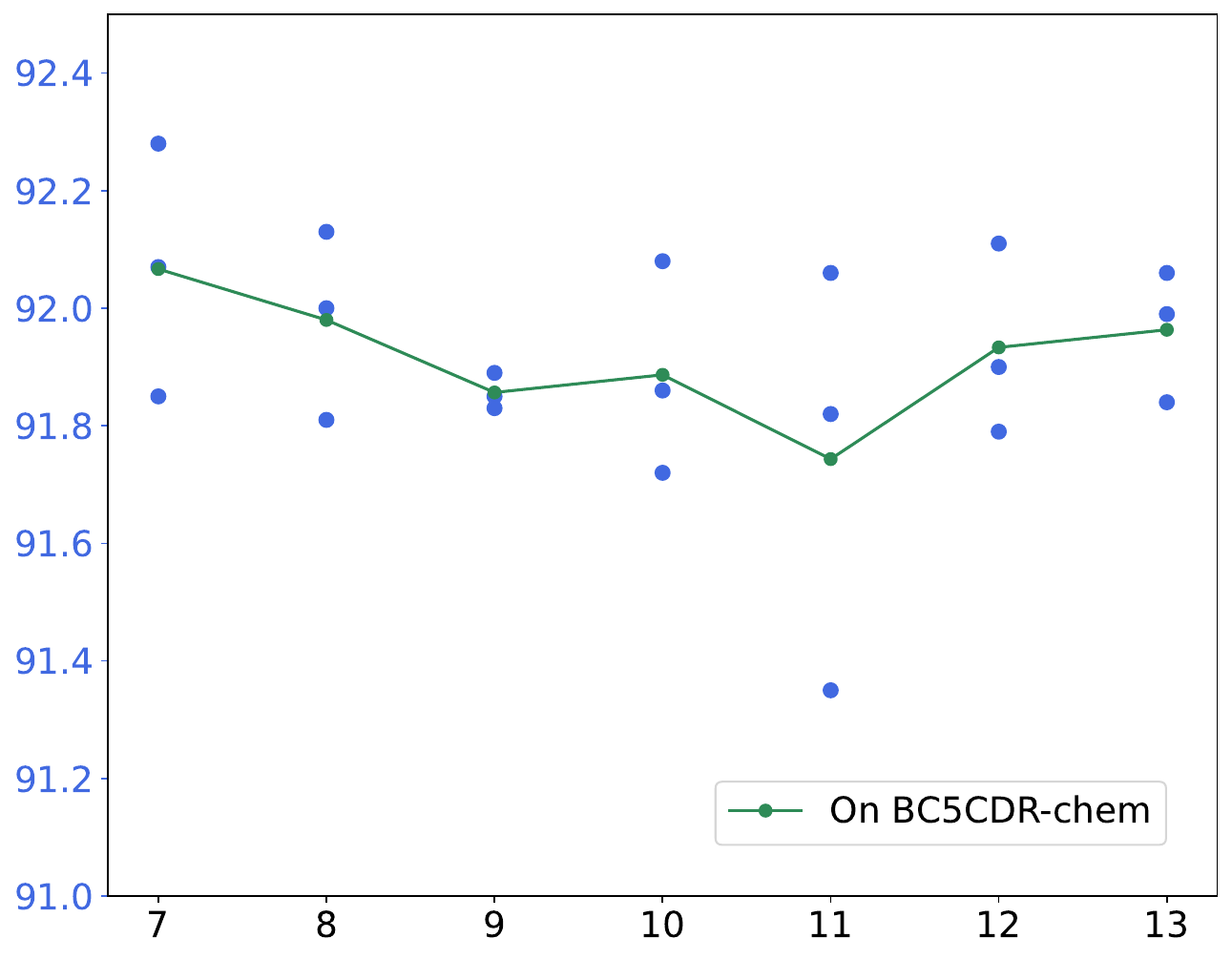}
        \caption{}
    \end{subfigure}
    \caption{The impact of sampling a different number of entities at each iteration in the dictionary refinement algorithm on the model's performance. The value x in x-axis represents sampled $2^x$ entities from KB in each iteration, and the y-axis represents the F1-score of the DMNER model. (a) and (b) represent the experiments conducted on the NCBI and BC5CDR-chem datasets, respectively.}
    \label{fig:aba4}
\end{figure}

\subsection{Knowledge Base}
A knowledge base is a collection of <entity-entity category> pairs. In this paper, the knowledge base has two data sources: the CTD (Comparative Toxicogenomics Database) and the UMLS  (Unified Medical Language System) knowledge base. The CTD knowledge base provides five types of entities, while the UMLS knowledge base is larger in scale. As of UMLS 2023, it contains 3,313,382 concepts and 15,718,215 concept names derived from over 150 controlled vocabularies. The CTD knowledge base provides download links for each type of entity, and the categories of the UMLS data are referenced to the UMLS semantic network tree.
\footnote{\url{https://www.nlm.nih.gov/research/umls/knowledge_sources/metathesaurus/release/statistics.html}}

\subsection{Dictionary refinement parameter}
This section focuses on parameter selection in the dictionary refinement algorithm, specifically the threshold $t$. In Figure \ref{fig:aba3}, we conducted experiments on the NCBI dataset and BC5CDR dataset using five different values of $t$, ranging from 0 to 4. 
In each setting, we experiment three times and show the averge score.
Upon careful observation, we found that when $t$ is set to 0 or 1, there is a slight decrease in the performance of the DMNER model. As $t$ increases beyond 1, there is no significant change in the overall performance, with some fluctuations observed. Based on these findings, we decided to set $t=2$ uniformly when performing dictionary refinement.

In Figure \ref{fig:aba4}, the impact of the number of entities sampled at each iteration in the dictionary refinement algorithm on the model's performance is investigated. For each setting, three experiments are conducted, and the average value is taken. 
To ensure that the number of entities encountered in the knowledge base remains consistent across different sampling sizes, we set the product of the number of iterations and the sampling quantity to be the same for each experiment. 
The experimental results indicate that the number of entities sampled does not have a significant impact on the performance of DMNER.

\label{sec:appendix}

\end{document}